\documentclass{fairmeta} %

\usepackage{url}
\usepackage{enumitem}
\usepackage{algorithm}
\usepackage{algorithmic}
\usepackage{amssymb}
\usepackage{amsmath}
\usepackage{array}
\usepackage{subcaption} %

\usepackage{amsmath,amsfonts, bm}
\usepackage{amssymb} 

\newcommand{\ie}{\textit{i}.\textit{e}.}

\def\eqref#1{equation~\ref{#1}}

\def\1{\bm{1}}

\DeclareMathAlphabet{\mathsfit}{\encodingdefault}{\sfdefault}{m}{sl}
\SetMathAlphabet{\mathsfit}{bold}{\encodingdefault}{\sfdefault}{bx}{n}

\newcommand{\rev}[1]{#1}
\newenvironment{revblock}{}{}
\newcommand{\revcolor}{}

\AtBeginEnvironment{tabular}{\small}
\AtBeginEnvironment{tabular*}{\small}

\title{The View from Nowhere: Population-Centered Agent Evolution}

\author[1]{Shuo Yan}
\author[1]{Ningning Hou}
\author[2]{Haoyu He}
\author[1]{Haochen Pei}
\author[1]{Shang Gao}

\affiliation[1]{Macquarie University}
\affiliation[2]{Monash University}

\abstract{%
Self-improving AI systems can revise the designs that govern their own behavior. Yet existing approaches center improvement on selected parent agents and their evolutionary experience, leaving the broader population largely underused. We identify a substantial gap between population coverage (\ie, the tasks solved by at least one evolved variant) and the performance of the best individual agent. This gap shows that useful capabilities and diagnostic evidence emerge across the population without being consolidated into its leading agents.
These observations motivate the \textbf{view from nowhere}, a principle for agent evolution: an agent should be improved by interpreting its behavior through population-wide patterns of success, failure, and prior self-modification. We instantiate this principle as \textbf{Population-Centered Evolution (PCE)}. PCE distils task records and modification histories from the population into a population block, which captures recurring failures, complementary successes, and the effects of previously attempted modifications to guide improvements to a selected parent’s harness.
Under matched models, initial harnesses, and evolution schedules, PCE achieves 72.1\% success on SWE-bench Verified and 96.7\% on Polyglot, improving over a strong group-evolution baseline by 3.2 and 5.9 percentage points. \rev{Across independent runs}, PCE reduces the gap between population coverage and best-agent performance by \rev{30\%} on SWE-bench and \rev{all but} closes it on Polyglot. These results suggest that treating the population as evidence, not merely as candidates for selection, provides a complementary principle for agent evolution.%
}

\metadata[Code]{\url{https://github.com/YanShuo1992/PCE}}
\metadata[Website]{\url{https://yanshuo1992.github.io/PCE/}}

\hypersetup{
  pdftitle={The View from Nowhere: Population-Centered Agent Evolution},
  pdfauthor={Shuo Yan, Ningning Hou, Haoyu He, Haochen Pei, Shang Gao}
}

\newcommand{\onelogo}[2][0.55cm]{\raisebox{-0.5\height}{\includegraphics[height=#1]{logos/#2}}}
\renewcommand{\titlelogos}{%
  \onelogo[0.75cm]{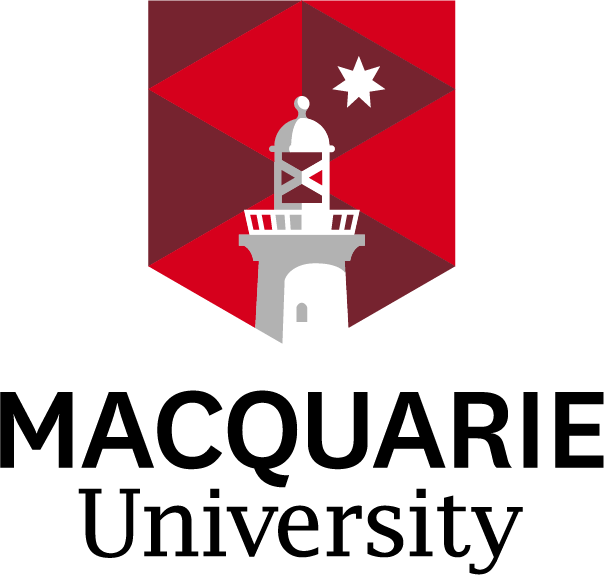}\;\onelogo[0.7cm]{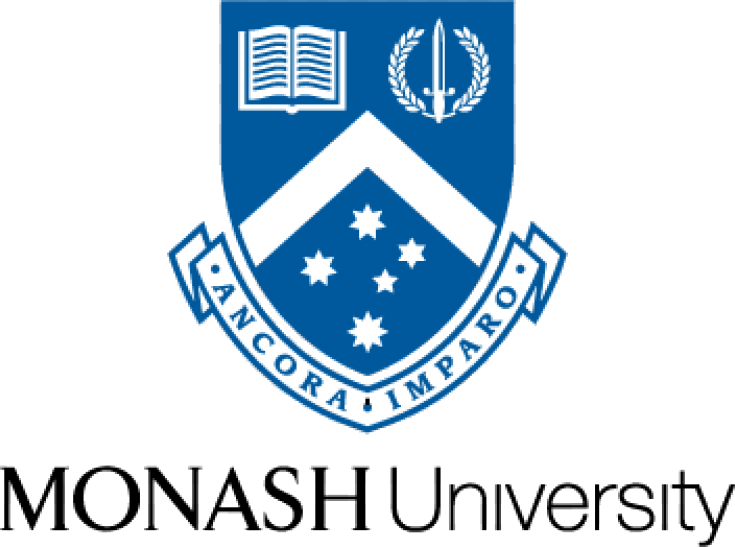}%
}

\begin{document}
\maketitle

\section{Introduction}
\label{sec:intro}

Self-improving AI aims to develop systems that can revise the designs governing their own behavior \citep{good1966speculations,schmidhuber2007godel}. Evolutionary coding agents pursue this goal by modifying their software harnesses while keeping the parameters of their foundation models fixed \citep{zhang2025dgm,weng2026gea,robeyns2025sica,wang2025hgm}. These harnesses determine how an agent uses the model, tools, feedback, and its environment to solve a task. Through repeated cycles of self-modification and evaluation, these systems generate a growing population of agent variants. The goal is not merely to produce diverse variants, but to turn useful discoveries made during evolution into progressively stronger agents. How to consolidate these discoveries into improvements that persist in leading agents is a central challenge for agent evolution.

Existing approaches address this challenge through parent selection and experience reuse. The Darwin G\"odel Machine (DGM) preserves an archive of agent variants from which future evolution can branch \citep{zhang2025dgm}. The Huxley-G\"odel Machine (HGM) uses the performance of an agent and its descendants to estimate the potential of continuing that lineage \citep{wang2025hgm}. Group-Evolving Agents (GEA) selects a small group of parents and allows them to share task records and past self-modifications when proposing improvements \citep{weng2026gea}. Despite these differences, the evidence used to propose a modification remains largely \emph{parent-local}: it comes from the selected parent or parent group rather than from a systematic comparison across the full population. Other variants may remain available for future selection, but their task-level successes, failures, and modification histories do not directly inform the current diagnosis.

\begin{figure}[t]
\centering
\includegraphics[width=0.228\linewidth]{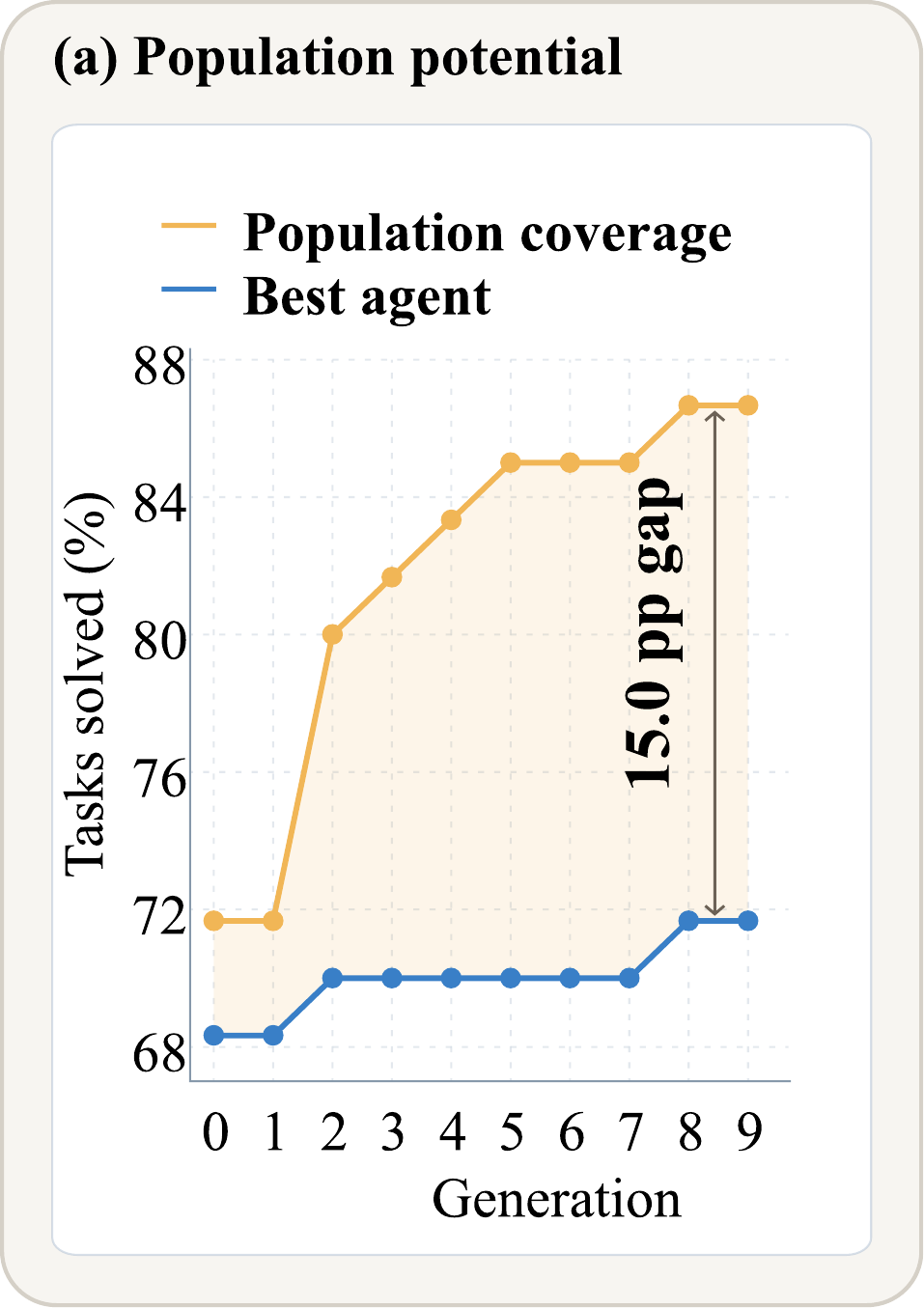}\hfill
\includegraphics[width=0.379\linewidth]{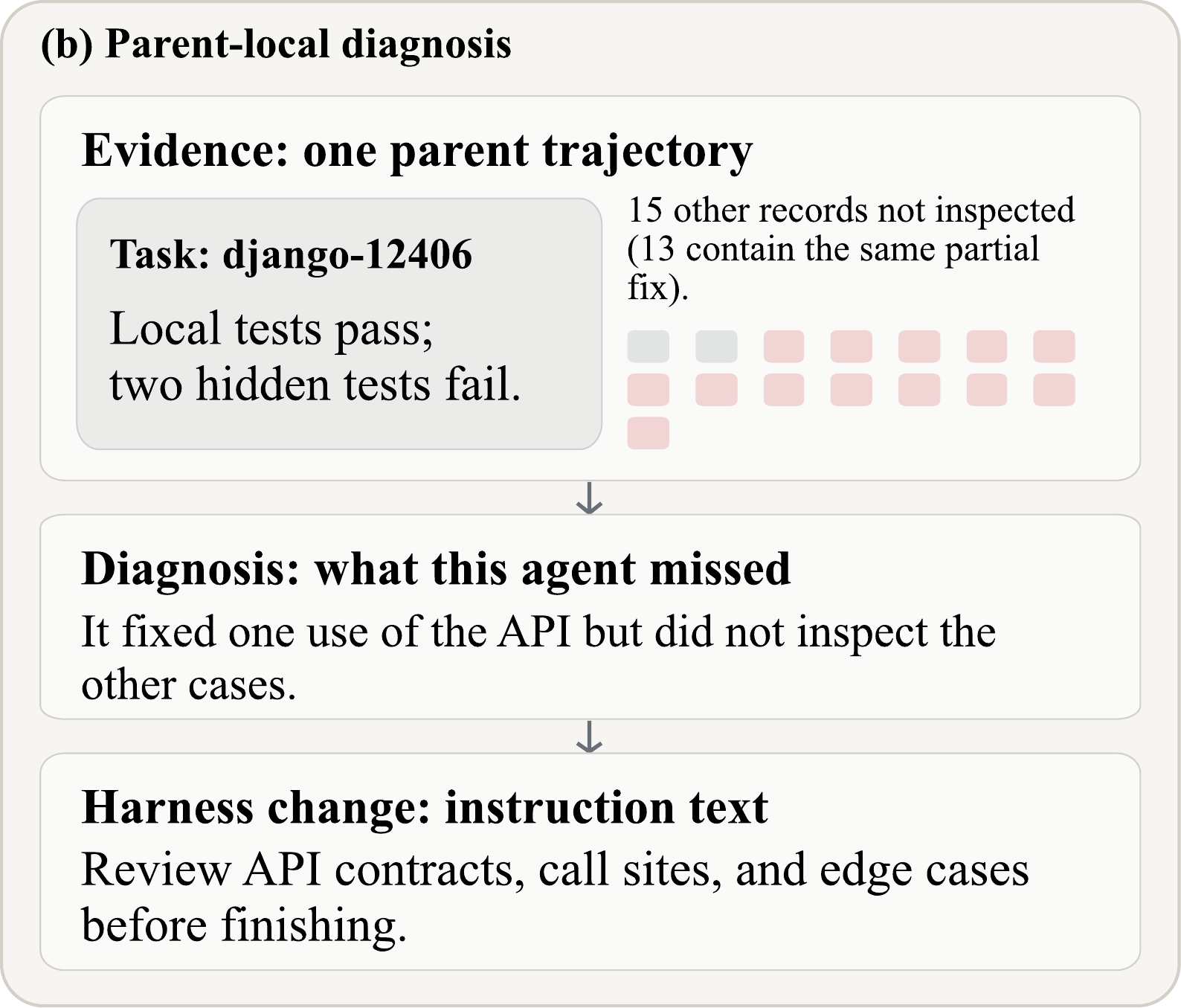}\hfill
\includegraphics[width=0.379\linewidth]{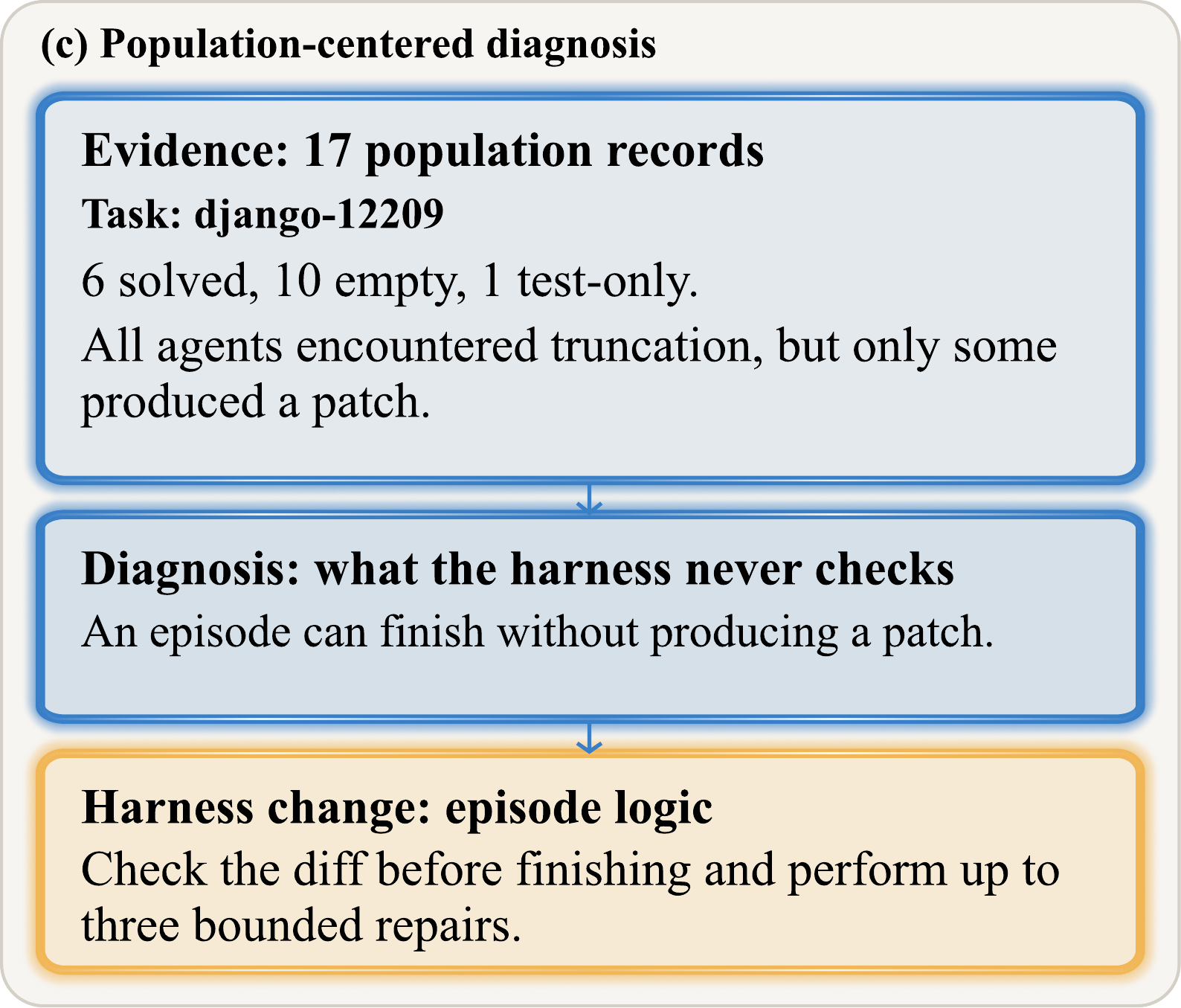}
\caption{Population potential and task-level evidence. (a) The union of tasks solved by the GEA population exceeds the performance of its best individual agent by 15.0 percentage points (9 tasks in Held-in 60). (b) Parent-local diagnosis uses one parent trajectory to identify missed API cases and add verification instructions, while the other 15 population records are not inspected. (c) Population-centered diagnosis uses patterns across 17 records to identify a missing patch check and add bounded repair logic.}
\label{fig:motivation-intro}
\vspace{-0.5cm}
\end{figure}

Yet prior work has not systematically measured the gap between what an evolutionary population collectively solves and what its best individual agent achieves, or examined how this gap can guide improvement. We define \emph{population coverage} as the proportion of tasks solved by at least one evolved variant. As shown in Figure~\ref{fig:motivation-intro}a, in one of our GEA runs the evolved population covers 9 more of the 60 held-in tasks than its best individual agent, corresponding to a 15.0 percentage-point gap. This gap shows that successful outcomes remain distributed across the population rather than being fully consolidated into its leading agent. The associated task records allow us to compare variants that succeed and fail on the same task. The population should therefore be treated not only as a pool for parent selection, but also as comparative evidence for deciding how a selected parent should improve. 

These observations motivate the \emph{view from nowhere} \citep{nagel1986view} as a principle for agent evolution. It holds that decisions about how to improve an agent should be grounded in patterns of success, failure, and prior self-modification across the population. For a given task, population outcomes show whether a failure is isolated or recurring, execution records reveal how successful and unsuccessful variants differ, and modification histories indicate which remedies have already been attempted. Together, these comparisons provide a broader basis for diagnosing a selected parent and guiding its improvement.

We instantiate this principle through \textbf{Population-Centered Evolution (PCE)}. For a selected task, a distillation agent compares task records and modification histories across all available variants and organizes its findings into a population block. The block contains three types of information. Outcome evidence measures how frequently different results occur across the population. An observation identifies a cross-agent pattern that may explain why variants behave differently. Candidate files indicate which parts of the harness may be related to that pattern. Together, these fields establish what recurs across the population, suggest a diagnostic hypothesis, and indicate where to investigate it. Diagnosis then checks this hypothesis against the selected parent’s execution records and source code before proposing a harness modification, as shown in Figure~\ref{fig:overview}b.

Across SWE-bench Verified and Polyglot, PCE improves agent performance under matched models, initial harnesses, and evolution schedules. Agents evolved by PCE achieve 72.1\% success on 200 SWE-bench tasks and 96.7\% on 225 Polyglot tasks, exceeding GEA by 3.2 and 5.9 percentage points, respectively. \rev{Across independent runs, }PCE reduces the gap between population coverage and best-agent performance by \rev{30\%} on SWE-bench and \rev{all but} closes it on Polyglot. Its advantage also persists on held-out tasks. Across the full evolutionary process, PCE uses 36\% fewer tokens on SWE-bench and 59\% fewer tokens on Polyglot than GEA. PCE therefore achieves stronger agents with lower inference expenditure, showing that population-wide evidence enables more effective use of the experience accumulated during evolution. 

In summary, we make three contributions:
\begin{enumerate}
\item We identify and quantify a previously overlooked gap between population coverage and best-agent performance, and we formulate the \emph{view from nowhere} as a principle for capturing useful information across the population to guide agent evolution.
\item We introduce PCE, which distils population-wide experience into a population block that guides improvements to a selected parent’s harness.
\item PCE improves performance and generalization, narrows the population-coverage gap, \rev{consolidates more population-wide successes into individual agents,} and achieves these gains with lower total token consumption than baselines.
\end{enumerate}

\section{Related Work}
\label{sec:related}

\paragraph{Learning from execution experience.}
Agents use execution experience at several levels. Reflection revises outputs or memories using execution or self-generated feedback \citep{shinn2023reflexion,madaan2023selfrefine}, while memory systems, skill libraries, and workflow memories retain useful experience for future tasks \citep{park2023generative,packer2023memgpt,xu2025amem,wang2023voyager,wang2024awm}. ExpeL goes beyond storing individual records by extracting natural-language insights from accumulated task experience \citep{zhao2024expel}. Experience can also support parameter updates through self-training on generated responses and filtered trajectories \citep{zelikman2022star,chen2024spin,gulcehre2023rest,yuan2024selfrewarding}. Our setting concerns design-level improvement while keeping the foundation model fixed. Self-Harness identifies model-specific failure patterns and validates harness edits with regression tests \citep{zhang2026selfharness}, while HarnessEvolve aligns failed executions with reference trajectories and clusters error signals to identify systematic failures \citep{jiang2026harnessevolve}. PCE differs in its unit of analysis. It compares how a population of evolving harness variants handles the same task and relates their outcomes to previous self-modifications. This population-wide comparison provides evidence for diagnosing and modifying the current parent.

\paragraph{Population-based agent design and evolution.}
Automated agent design methods search over agentic modules, workflows, prompts, and code \citep{hu2025adas,zhang2024aflow,wang2025evoagentx,fernando2023promptbreeder,zelikman2023stop,yin2025godel,robeyns2025sica}. As this search produces more variants, a central problem is how to preserve and reuse discoveries made during evolution. Quality-diversity and open-endedness research retains diverse solutions as stepping stones for continued exploration \citep{lehman2011novelty,mouret2015mapelites,pugh2016qd,chatzilygeroudis2021qd,stanley2015greatness,stanley2017openendedness,clune2019aiga,hughes2024openendedness,faldor2025omniepic,wang2019poet}. In coding-agent evolution, DGM maintains an archive from which future evolution can branch \citep{zhang2025dgm}, while HGM estimates the potential of continuing a lineage from the performance of an agent and its descendants \citep{wang2025hgm}. GEA shares task records and self-modification patches among selected parents \citep{weng2026gea}, and HarnessBank recombines harnesses drawn from a pool \citep{luo2026harnessbank}. These approaches use the population to preserve diversity, select parents, share parent experience, or recombine designs. PCE gives the population a different role. It treats all available variants as comparative evidence for understanding recurring failures, contrasting outcomes, and guiding improvement of a selected parent.

\section{Population-Centered Evolution}
\label{sec:method}
\begin{figure}[t]
\centering
\includegraphics[width=\linewidth]{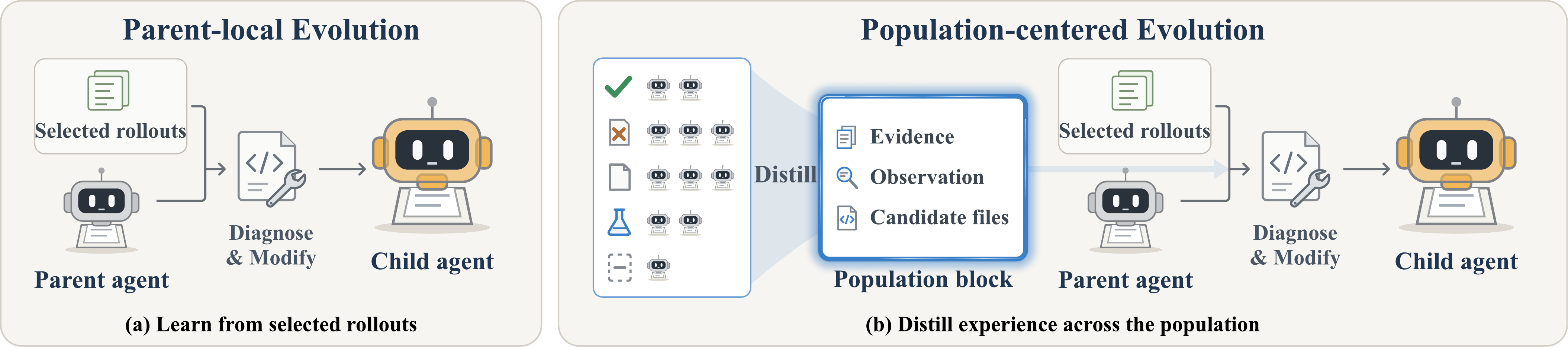}

\caption{
Parent-local and population-centered evolution. (a) Parent-local evolution diagnoses and modifies an agent using execution records associated with the selected parent. (b) Population-centered evolution distils task records and modification histories across the population into a population block. The block summarizes outcome evidence, identifies cross-agent observations, and suggests candidate files for inspection. Diagnosis combines this block with the selected parent’s records and source code to propose a harness modification.
}
\label{fig:overview}
\end{figure}

PCE operates population-centered evolution through two stages, as shown in Figure~\ref{fig:overview}. Population distillation compares task records and modification histories across evaluated agent variants and summarizes the findings in a population block. Diagnosis then checks this population-level account against a selected parent's records and source code before proposing a harness modification.

We note that PCE distinguishes the variants used as evidence from those eligible for parent selection. It maintains a population $\mathcal{V}$ containing all evaluated agent variants whose task records are available, and an archive $\mathcal{A}\subseteq\mathcal{V}$ containing the variants eligible to become parents. Each variant is associated with its harness, task records, and history of self-modification. A variant can therefore inform future improvements even when it is not selected as a parent. 

At each evolution step, PCE selects parent agents from $\mathcal{A}$ and chooses a task for diagnosing each parent. Population distillation examines the records available in $\mathcal{V}$ for that task, after which diagnosis proposes and implements a modification to the selected parent's harness. The resulting child is evaluated and added to $\mathcal{V}$. Children satisfying the archive criteria also enter $\mathcal{A}$ for future parent selection.

\subsection{Population Distillation}
\label{sec:method:distill}

For the task selected to diagnose a parent, population distillation considers all available task records in $\mathcal{V}$, including the initial agent's record when available. Each record contains an execution log, a submitted task patch, and an evaluation outcome. The corresponding harness can be reconstructed from the initial source and the variant's chain of self-modifications. These materials allow distillation to compare not only which variants succeed or fail, but also how their execution and harness designs differ.

Population distillation combines programmatic measurement with agent-based analysis. A program first counts the outcomes across all available records. The distillation agent then inspects relevant execution logs, task patches, harness source, and earlier self-modifications. It compares successful and unsuccessful variants, identifies recurring outcomes, and examines whether previous harness changes resolved or preserved the observed behavior. Since the full records can be large, the agent chooses which records and source files to inspect within its analysis budget.

The resulting population block contains three fields. \textit{Outcome evidence} reports the program-computed distribution of task outcomes across the population. A \textit{cross-agent observation} summarizes a pattern found by comparing variants and provides a hypothesis about why their outcomes differ. \textit{Candidate files} identify parts of the harness that may implement the relevant behavior and should be examined during diagnosis. This field is left empty when distillation cannot identify a plausible location. Separating measured outcomes from their interpretation preserves a factual population summary while giving diagnosis a focused starting point without prescribing the modification. 
Appendix~\ref{app:details} gives the output contract and the distillation budget.

\subsection{Diagnosis and Harness Modification}
\label{sec:method:diagnosis}

The population block provides a diagnostic hypothesis rather than prescribing a modification. Diagnosis combines this block with the selected parent's source code and task records to determine whether the population-level pattern applies to the current harness. It examines the candidate files when provided, but may inspect other parts of the harness as needed. A mechanism associated with successful variants may be absent from the parent, while an earlier modification may already be present but remain insufficient. Diagnosis uses the parent's execution behavior and implementation to distinguish between these possibilities.

Based on this assessment, diagnosis proposes one harness modification and describes how it should be implemented. The self-modification step applies the proposal to a copy of the parent's harness, producing a child agent. The modification may change the agent's prompts, tools, or execution workflow, while the foundation model and inference limits remain fixed. The child is then passed to the evaluation stage of the evolution loop.

The SWE-bench task django-12209 in Figure~\ref{fig:motivation-intro}c illustrates the full process. Across 17 task records, the outcome evidence reported 6 solved patches, 10 empty patches, and 1 test-only patch. Population distillation found that truncation occurred under different outcomes and that increasing the retry budget had not consistently prevented empty patches. The observation and candidate files directed diagnosis toward the parent's episode-completion logic. Inspection then showed that the episode could finish without checking whether a patch had been produced. The child harness then added this check and performed bounded repairs when the patch remained empty. This example shows how population-wide patterns are converted into a hypothesis, validated against the selected parent, and implemented as a concrete harness change.

\subsection{Evolution Loop}
\label{sec:method:loop}

Algorithm~\ref{alg:loop} specifies the complete PCE evolution loop. The first generation produces two children from the initial agent through separate single-parent diagnoses. Later generations select two parents \rev{in our setting} from $\mathcal{A}$ and sample one eligible diagnostic task for each. For a selected task, population distillation runs only when at least three records are available in $\mathcal{V}$. Otherwise, or when no usable observation is produced, diagnosis falls back to the parent-local procedure.

Children that produce a self-modification patch are evaluated first on an initial subset and then on the remaining evolution tasks if they meet the evaluation threshold. Every evaluated child joins $\mathcal{V}$, while children satisfying the retention criteria also join $\mathcal{A}$. Appendix~\ref{app:details} provides the task-sampling rules, record ordering, evaluation thresholds, and archive-update criteria.

\begin{algorithm}[t]
\caption{PCE evolution loop}
\label{alg:loop}
\begin{algorithmic}[1]
\REQUIRE Initial agent $h_0$ with evaluation records; number of generations $T$
\STATE Initialize population and archive $\mathcal{V}=\mathcal{A}=\{h_0\}$
\FOR{generation $t=0,\ldots,T-1$}
  \STATE Select parents $P_t=\{h_1,\ldots,h_k\}$ from $\mathcal{A}$; use $P_t=\{h_0\}$ if $t=0$
  \STATE Each parent $h_i\in P_t$ samples a task $\tau_i$ from its eligible tasks
  \STATE Form inputs $L_i$: the task record of $h_i$, followed by the records of the other parents in $P_t$
  \FOR{each $h_i\in P_t$ \textbf{in parallel}}
    \STATE Collect records $R_i$ on task $\tau_i$ from members of $\mathcal{V}$
    \STATE Set population block $B_i=\varnothing$
    \IF{$|R_i|\geq 3$}
      \STATE $B_i\leftarrow\textsc{Distill}(R_i,\text{patch histories},h_0)$
      \STATE Set $B_i=\varnothing$ if no usable observation is returned
    \ENDIF
    \STATE Diagnose the harness of $h_i$ using $L_i$ and any available $B_i$
    \STATE Implement the proposal in $h_i$ to produce child $h_{c,i}$
    \IF{a self-modification patch is produced}
      \STATE Evaluate $h_{c,i}$ on the initial task subset, then remaining tasks if the threshold is met
      \STATE Retain its execution logs, task patches, and evaluation outcomes
    \ENDIF
  \ENDFOR
  \STATE Add evaluated children to $\mathcal{V}$ and those meeting archive retention criteria to $\mathcal{A}$
\ENDFOR
\RETURN Archive $\mathcal{A}$ and population $\mathcal{V}$ with accumulated records
\end{algorithmic}
\end{algorithm}

\section{Experiments Setup}
\label{sec:experiments}

\subsection{Benchmarks and Data Splits}
\label{sec:exp:benchmarks}

We evaluate PCE on SWE-bench Verified and Polyglot \citep{jimenez2024swebench,openai2024verified,gauthier2024polyglot}. SWE-bench Verified evaluates repository-level issue resolution, while Polyglot contains 225 coding tasks across six programming languages. Following DGM, we use 200 SWE-bench Verified tasks and all 225 Polyglot tasks \citep{zhang2025dgm}. For each benchmark, 60 tasks are held in for evolution. The remaining 140 SWE-bench tasks and 165 Polyglot tasks are held out for final evaluation, forming Overall 200 and Overall 225. We also evaluate on an additional set of 60 SWE-bench tasks spanning 12 repositories.

\subsection{Baselines and Settings}
\label{sec:exp:conditions}
\label{sec:exp:setup}

We compare PCE with GEA, the closest group-evolution baseline \citep{weng2026gea}. To isolate the effect of population-centered evolution, both methods use the same foundation models, initial harness, parent-selection procedure, diagnosis prompt, modifiable files, and evaluation protocol. Each method runs for ten generations and produces two children per generation. \rev{We conduct two independent runs of each method on SWE-bench and three on Polyglot. The Best column of Table~\ref{tab:final}, Table~\ref{tab:cohort}, Figures~\ref{fig:main}a and~\ref{fig:main}b, and Tables~\ref{tab:runs} and~\ref{tab:runs-polyglot} report all runs. Unless stated otherwise, all other results come from the main run of each benchmark, which is Run~1 in Tables~\ref{tab:runs} and~\ref{tab:runs-polyglot}.} DeepSeek Flash performs task execution and self-modification. DeepSeek Pro performs diagnosis for both methods and population distillation for PCE. We also evaluate the shared initial agent as a reference.

\subsection{Evaluation Protocol}
\label{sec:exp:evaluation}

During evolution, each child is first evaluated on a screening set of 10 tasks. Children meeting the advancement threshold are then evaluated on the remaining 50 held-in tasks. Performance across all 60 held-in tasks defines the search score, and the highest-scoring evolved agent is selected for final evaluation.

After evolution, we evaluate the selected agent without further self-modification. For each method, the selected agent and the initial agent are independently evaluated 5 times on the 60 held-in tasks. We report the mean success rate and sample standard deviation. The additional SWE-bench set of 60 tasks is also evaluated 5 times. SWE-bench Held-out 140 and Polyglot Held-out 165 are each evaluated once. Overall scores are task-count-weighted averages of the mean held-in score and the corresponding held-out score. Appendices~\ref{app:details} and~\ref{app:results} provide the task splits, model versions, advancement thresholds, and the results of all runs.

\section{Results and Analysis}
\label{sec:results}

\subsection{Agent Performance and Generalization}
\label{sec:res:main}
\label{sec:res:transfer}

\noindent\textbf{RQ1: Does PCE evolve stronger agents that generalize beyond the search evaluation?}

\textbf{PCE outperforms GEA on both benchmarks under matched experimental settings.}
Table~\ref{tab:final} reports the performance of the best agents evolved within 10 generations. Both GEA and PCE improve substantially over the shared initial agent. PCE reaches an Overall success rate of 72.1\% on SWE-bench and 96.7\% on Polyglot, exceeding GEA by 3.2 and 5.9 percentage points, respectively. The Polyglot result is particularly notable because PCE continues to improve performance even when GEA already solves more than 90\% of the tasks \rev{in the Overall evaluation}.

\begin{figure}[t]
\centering
\begin{subfigure}[b]{0.32\linewidth}
\includegraphics[width=\linewidth]{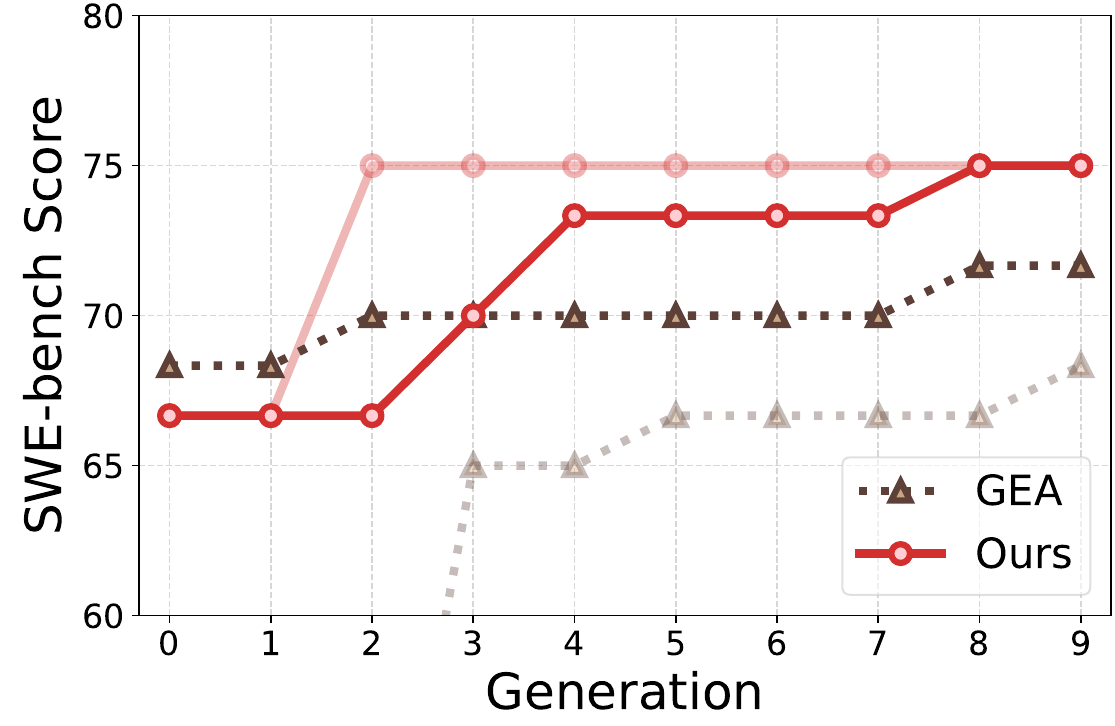}
\caption{SWE-bench search}
\end{subfigure}\hfill
\begin{subfigure}[b]{0.32\linewidth}
\includegraphics[width=\linewidth]{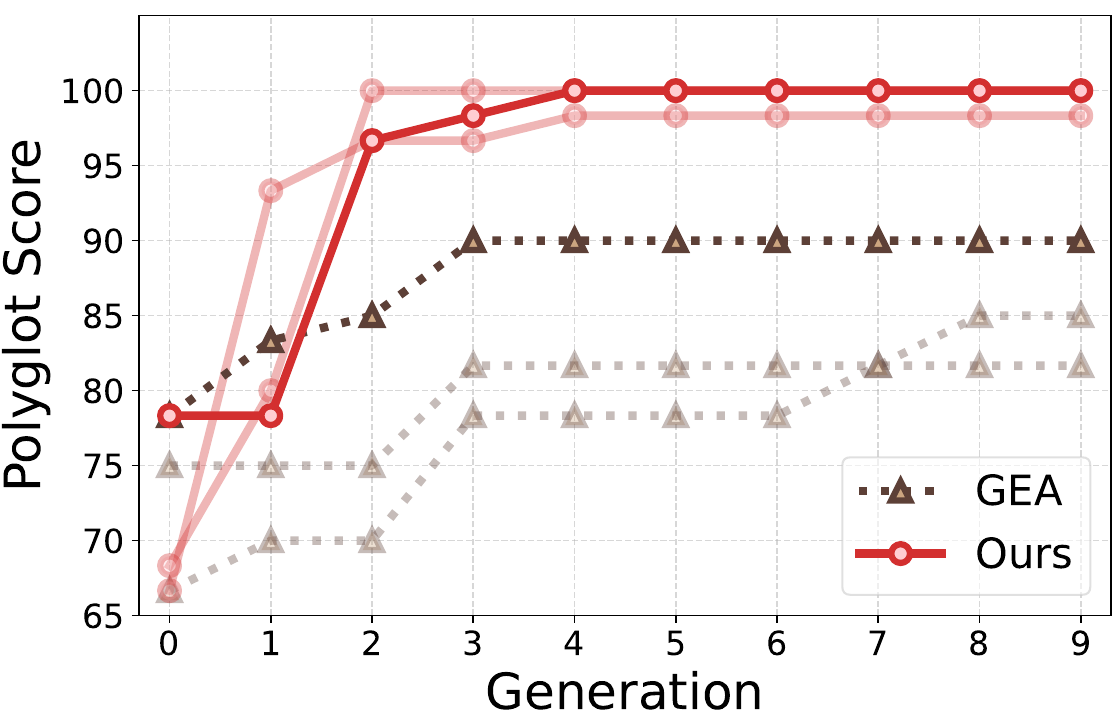}
\caption{Polyglot search}
\end{subfigure}\hfill
\begin{subfigure}[b]{0.32\linewidth}
\includegraphics[width=\linewidth]{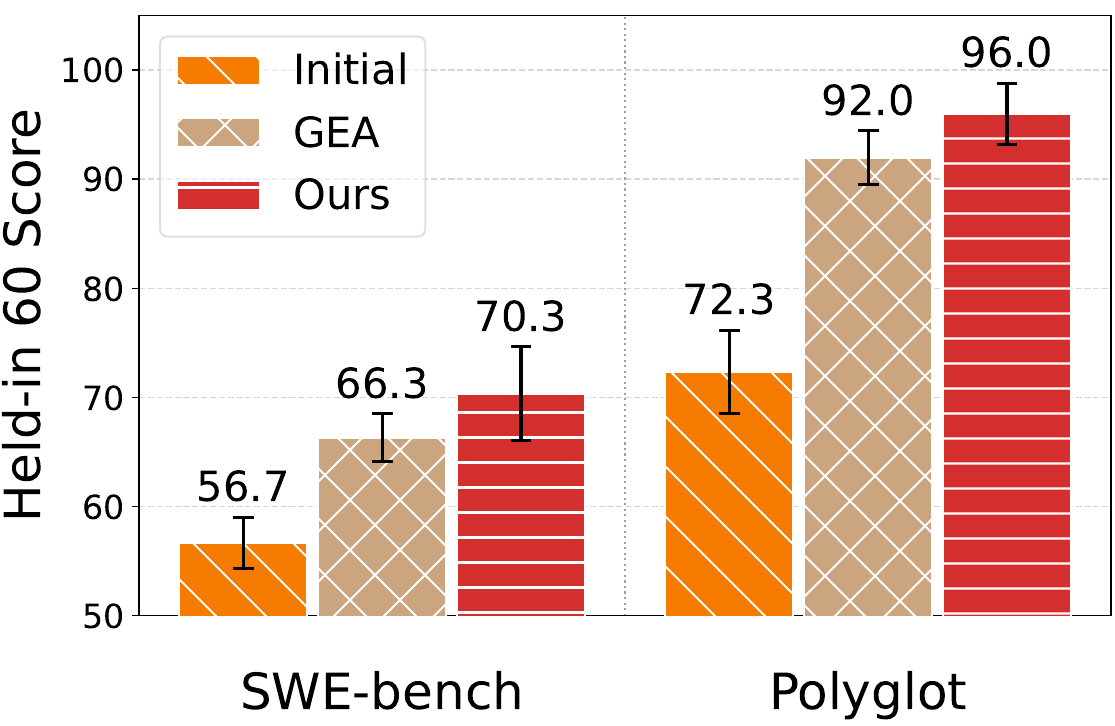}
\caption{Repeated evaluations}
\end{subfigure}

\caption{Search scores and repeated-evaluation results on Held-in 60. (a, b) Best-so-far search score across the main ten-generation run, with two children produced in each generation. \rev{Lighter lines show the additional independent runs of each method, one on SWE-bench and two on Polyglot. In the SWE-bench panel, the early scores of the second GEA run lie below the axis range.}
\rev{(c)} Mean success rate of the shared initial agent and the selected GEA and PCE agents over five independent evaluations after evolution.}
\label{fig:main}
\end{figure}

The gains are not limited to a single search or evaluation. \rev{Figures~\ref{fig:main}a and~\ref{fig:main}b plot two SWE-bench runs and three Polyglot runs of each method. Every PCE run achieves a higher best search score than any GEA run on the same benchmark. The weakest PCE run still exceeds the strongest GEA run by 3.3 points on SWE-bench and 8.3 points on Polyglot, as Tables~\ref{tab:runs} and~\ref{tab:runs-polyglot} show. The advantage therefore does not rest on a single favorable run.} When the selected agents are independently evaluated 5 times on Held-in 60, PCE retains a mean advantage of 4.0 percentage points on both benchmarks (Figure~\ref{fig:main}c). PCE also outperforms GEA on all three held-out task sets (Appendix~\ref{app:results}, Table~\ref{tab:repeats}). Since held-out tasks constitute the majority of Overall 200 and Overall 225, these results show that the gains extend beyond the tasks used during evolution.

For context, Table~\ref{tab:final} also reports published results for DGM, GEA, OpenHands, and Aider under their original models and protocols. These values are reference points rather than controlled comparisons. Our main conclusion is based on the matched PCE and GEA runs.

\subsection{Population-Wide Improvement and Consolidation}
\label{sec:res:coverage}

\noindent\textbf{RQ2: Does PCE consolidate population-wide successes into stronger individual agents?}

\textbf{PCE improves not only the best agent, but also the broader set of leading variants.}
Table~\ref{tab:cohort} reports the highest search score and the minimum score among the top three and top five evolved variants. PCE improves all three measures on both benchmarks. On SWE-bench, even the fifth-ranked PCE variant performs comparably to GEA's best variant. On Polyglot, the fifth-ranked PCE variant outperforms GEA's best. The gains therefore extend beyond a single exceptional agent.

PCE also brings its leading agents closer to what the population collectively solves. Population coverage measures the fraction of tasks solved by at least one evaluated variant, while the coverage gap is the difference between this union and the best-agent score. On SWE-bench, PCE and GEA achieve similar population coverage, but PCE reduces the mean gap from 14.2 to 10.0 percentage points. On Polyglot, both populations cover all 60 held-in tasks \rev{in every run}. GEA retains a \rev{mean gap of 14.4 points, whereas PCE reduces it to 0.6 points. In two of the three runs, a single PCE agent solves the full task set, and in the remaining run the best agent misses one task}.

The reduction does not result from PCE exploring a substantially larger set of solvable tasks. Instead, with similar population coverage, PCE produces leading variants that capture more of the successful outcomes distributed across the population. This result directly supports the central motivation for using population-wide evidence during agent evolution.

\rev{PCE children also acquire more of what earlier population members solve. Table~\ref{tab:cohort} reports this as the consolidation rate, which Appendix~\ref{app:consolidation} defines and reports for every run. On SWE-bench, PCE achieves a consolidation rate of 35.6\%, compared with 33.1\% for GEA. On Polyglot, the rate is 72.9\% for PCE against 44.6\% for GEA, and the lowest PCE run exceeds the highest GEA run. Successes found by other population members therefore reach PCE's children more often, which provides direct evidence of cross-agent consolidation.}

\begin{table}[t]
\caption{Performance (\%). Best: highest Held-in 60 search score\rev{, mean $\pm$ sd over runs}. $\dagger$: reported by GEA \citep{weng2026gea}. $\ddagger$: model transfer. \rev{Settings} in Appendix~\ref{app:details}.}
\label{tab:final}
\centering
\setlength{\tabcolsep}{3pt}
\begin{tabular*}{\linewidth}{@{\hspace{8pt}\extracolsep{\fill}}llcccc@{\hspace{8pt}}}
\toprule
\textbf{Benchmark} & \textbf{Method} & \textbf{Model} & \textbf{Gen.} & \textbf{Best}$\uparrow$ & \textbf{Overall}$\uparrow$ \\
\midrule
\multirow{9}{*}{SWE-bench}
 & OpenHands$^\dagger$ & GPT-5 & N/A & N/A & 71.8 \\
\cmidrule(lr){2-6}
 & DGM & Claude 3.5 Sonnet & 80 & --- & 50.0 \\
 & DGM$^\ddagger$ & Claude 3.7 Sonnet & 80 & --- & 59.5 \\
 & DGM$^\ddagger$ & o3-mini & 80 & --- & 33.0 \\
 & DGM (GEA)$^\dagger$ & Haiku/Sonnet 4.5 & 60 & --- & 56.7 \\
 & GEA$^\dagger$ & Haiku/Sonnet 4.5 & 30 & --- & 71.0 \\
\cmidrule(lr){2-6}
 & Initial & DeepSeek-V4-Flash & 0 & N/A & 52.0 \\
 & GEA & DeepSeek-V4-Flash & 10 & \rev{70.0 $\pm$ 2.4} & 68.9 \\
 & \textbf{PCE} & DeepSeek-V4-Flash & 10 & \rev{\textbf{75.0} $\pm$ 0.0} & \textbf{72.1} \\
\midrule
\midrule
\addlinespace[2pt]
\multirow{9}{*}{Polyglot}
 & Aider$^\dagger$ & GPT-5 (high) & N/A & N/A & 52.0 \\
\cmidrule(lr){2-6}
 & DGM & o3-mini & 80 & --- & 30.7 \\
 & DGM$^\ddagger$ & Claude 3.5 Sonnet & 80 & --- & 33.3 \\
 & DGM$^\ddagger$ & Claude 3.7 Sonnet & 80 & --- & 36.8 \\
 & DGM (GEA)$^\dagger$ & Haiku/Sonnet 4.5 & 40 & --- & 68.3 \\
 & GEA$^\dagger$ & Haiku/Sonnet 4.5 & 20 & --- & 88.3 \\
\cmidrule(lr){2-6}
 & Initial & DeepSeek-V4.1-Flash & 0 & N/A & 65.1 \\
 & GEA & DeepSeek-V4.1-Flash & 10 & \rev{85.6 $\pm$ 4.2} & 90.8 \\
 & \textbf{PCE} & DeepSeek-V4.1-Flash & 10 & \rev{\textbf{99.4} $\pm$ 1.0} & \textbf{96.7} \\
\bottomrule
\end{tabular*}

\end{table}

\subsection{Value of Population-Wide Evidence}
\label{sec:res:process}

\noindent\textbf{\rev{RQ3: What in population-wide evidence drives the gains of PCE?}}

\rev{We conduct ablations on Polyglot under the same evolution and evaluation settings as the main run, and the variants differ only in the evidence that diagnosis receives. Besides parent-only distillation, member summaries condense the record of every population member on the task into a separate summary, so diagnosis sees the whole population but receives no population-level field. Three further variants keep outcome evidence, the cross-agent observation, or both, and the full population block of PCE adds candidate files. These controls separate the value of population-level evidence from the effect of an additional analysis step and from access to the records of the whole population.}

\textbf{Population-wide evidence improves performance beyond parent-only distillation.}
As shown in Table~\ref{tab:parentblock}, parent-only distillation achieves a best search score of 91.7\%, \rev{slightly above GEA's 90.0\% in the main run}. Both methods obtain the same mean success rate of 92.0\% across 5 evaluations of their selected agents. PCE instead reaches a perfect search score \rev{in the main run, 99.4\% on average over three runs,} and retains a mean success rate of 96.0\%, 4 percentage points above both alternatives.

The repeated evaluations further show that PCE solves more tasks consistently. All three methods solve nearly the full task set at least once, as reflected by their similar Union scores. PCE nevertheless achieves the highest Intersection, solving 88.3\% of the tasks in every evaluation, compared with 85.0\% for GEA and 78.3\% for parent-only distillation. \rev{The gains of PCE therefore come from the evidence that distillation supplies rather than from an additional analysis step alone.}

\begin{figure}[t]
\centering
\begin{subfigure}[b]{0.32\linewidth}
\includegraphics[width=\linewidth]{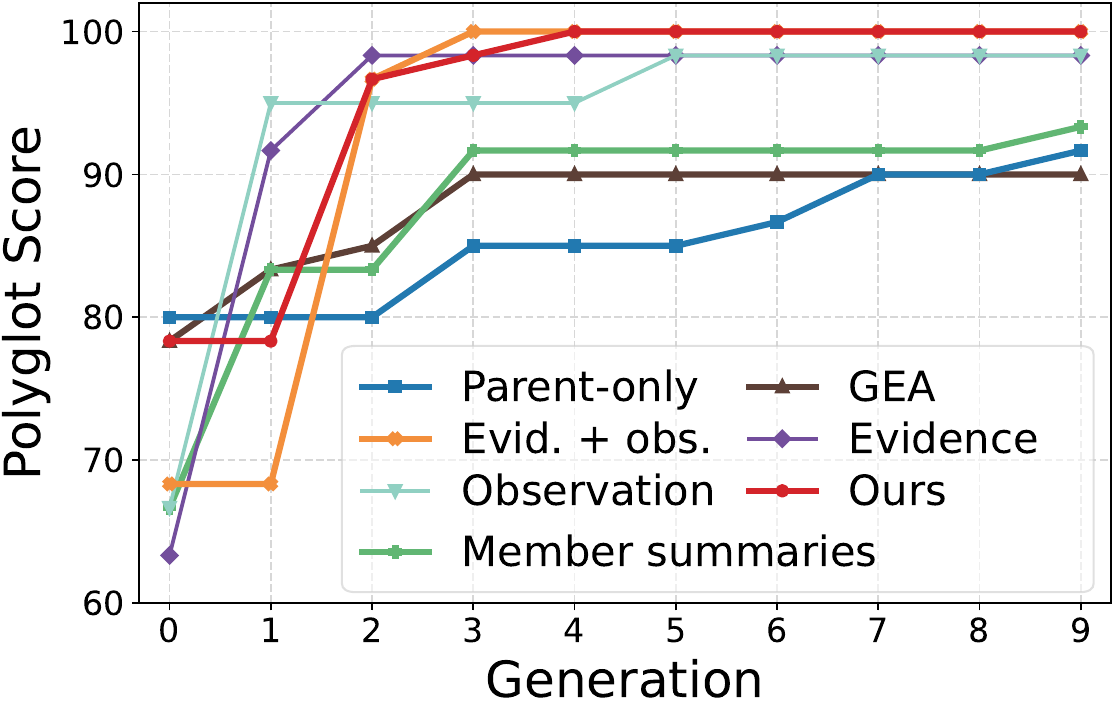}
\caption{\rev{Population block ablation}}
\end{subfigure}\hfill
\begin{subfigure}[b]{0.32\linewidth}
\includegraphics[width=\linewidth]{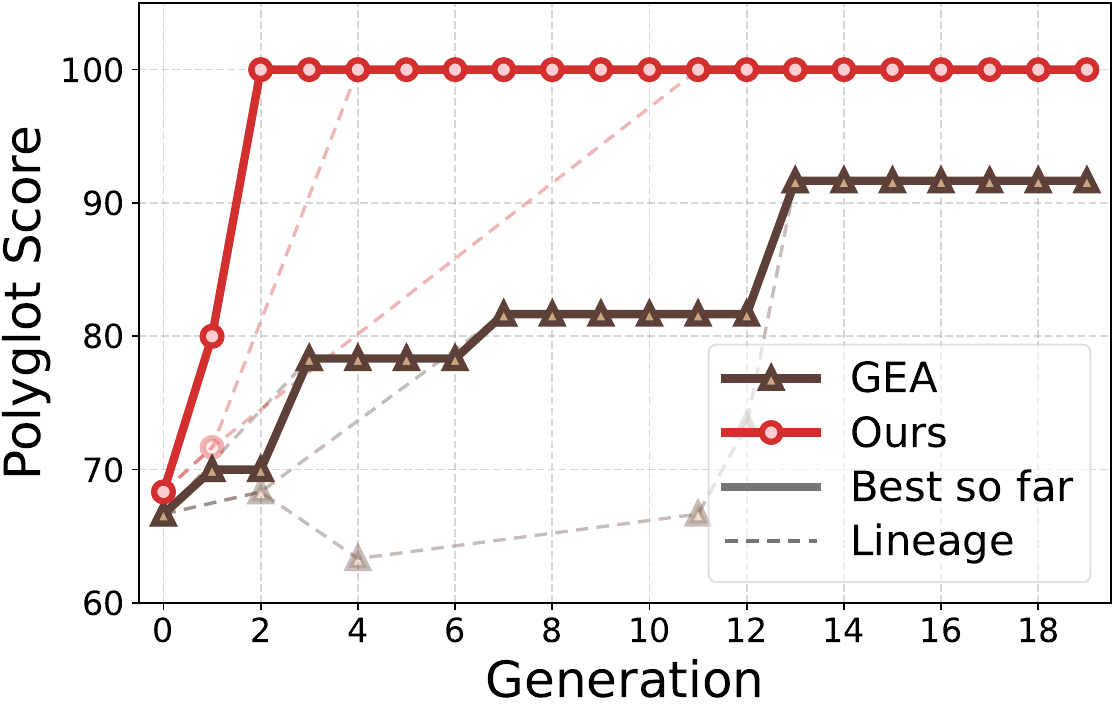}
\caption{\rev{Doubled generations}}
\end{subfigure}\hfill
\begin{subfigure}[b]{0.32\linewidth}
\includegraphics[width=\linewidth]{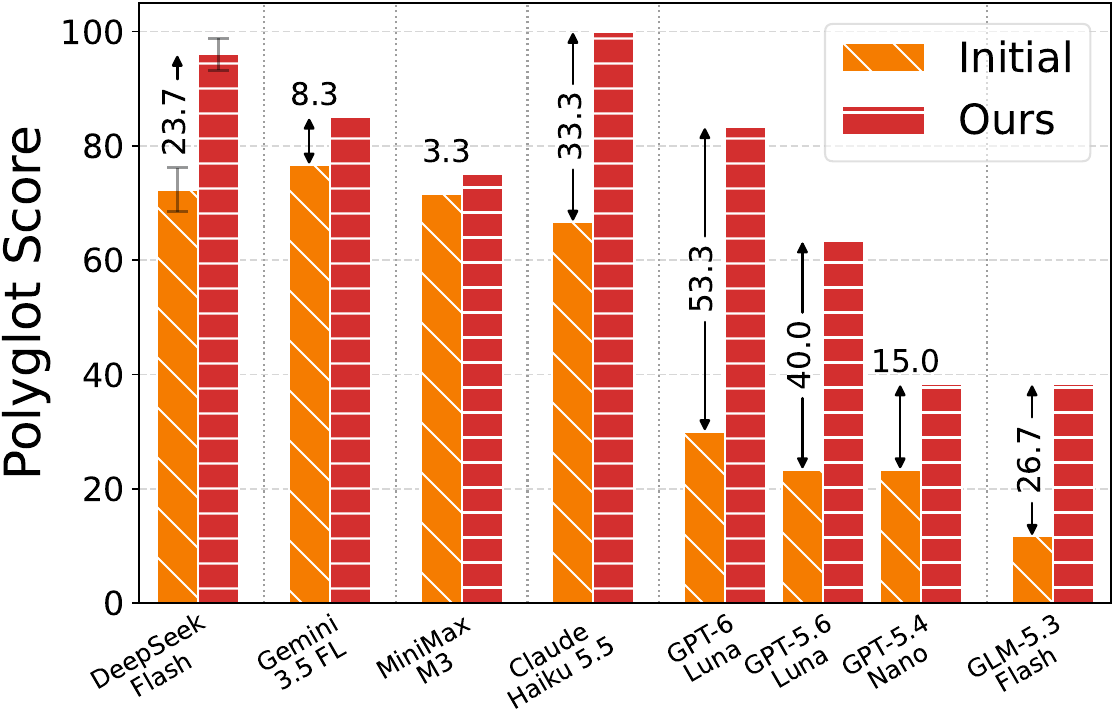}
\caption{\rev{Execution model transfer}}
\end{subfigure}
\caption{\rev{Results on Polyglot Held-in 60. (a) Best-so-far search score of the population block ablations. (b) Best-so-far search score and lineages of the best agents over twenty generations. (c) Success rate of the initial and PCE agents under eight execution models.}}
\label{fig:rq5}
\end{figure}

\begin{revblock}
\textbf{Population-level evidence, rather than more records or more analysis, drives the gains.}
The variants fall into two groups according to whether diagnosis receives population-level evidence, as Figure~\ref{fig:rq5}a and Table~\ref{tab:runs-polyglot} show. Whether it keeps outcome evidence, the cross-agent observation, or the full block, every variant with population-level evidence scores between 95.0\% and 98.3\% at generation~2 and ends between 98.3\% and 100\%. These variants end within one task of each other, so the fields provide the same population-level view rather than separate contributions. Every variant without it scores between 80.0\% and 85.0\% at generation~2 and ends between 90.0\% and 93.3\%, whether diagnosis sees only the parents or a summary of every member. More records therefore do not substitute for population-level evidence.

Fast convergence shows that the evidence behind the key modifications is already present in the early records of the population, and population-level evidence lets diagnosis find it. Without this evidence, the best score keeps rising in isolated steps throughout the run. Parent-only distillation and member summaries both still improve in the last generation, and GEA in Figure~\ref{fig:rq5}b still improves at generation~13. Each of these steps depends on a child that happens to act on the relevant evidence. The gains of PCE therefore come from population-level evidence rather than from more records or an additional analysis step.
\end{revblock}

\begin{table}[t]
\caption{Leading-variant performance and population coverage on Held-in 60 (\%)\rev{, mean $\pm$ sd over runs. Top-$k$ min. is the $k$-th highest search score, Union counts tasks solved by any child, Gap is Union minus Top-1, and Consolidation is defined in Appendix~\ref{app:consolidation}.}}
\label{tab:cohort}
\centering
\small
\setlength{\tabcolsep}{2pt}
\begin{tabular*}{\linewidth}{@{\extracolsep{\fill}}lcccccccc@{}}
\toprule
\textbf{Benchmark} & \textbf{Method} & \textbf{Top-1}$\uparrow$ & \textbf{Top-3 min.}$\uparrow$ & \textbf{Top-5 min.}$\uparrow$ & \textbf{Union} & \textbf{Gap}$\downarrow$ & \rev{\textbf{Consolidation}$\uparrow$} \\
\midrule
\multirow{2}{*}{SWE-bench} & GEA & 70.0 $\pm$ 2.4 & 66.7 $\pm$ 2.4 & 66.7 $\pm$ 2.4 & 84.2 $\pm$ 3.5 & 14.2 $\pm$ 1.2 & \rev{33.1 $\pm$ 10.7} \\
 & PCE & \textbf{75.0} $\pm$ 0.0 & \textbf{72.5} $\pm$ 1.2 & \textbf{70.8} $\pm$ 1.2 & 85.0 $\pm$ 0.0 & \textbf{10.0} $\pm$ 0.0 & \rev{\textbf{35.6} $\pm$ 9.0} \\
\midrule
\multirow{2}{*}{Polyglot} & GEA & \rev{85.6 $\pm$ 4.2} & \rev{81.1 $\pm$ 3.5} & \rev{78.3 $\pm$ 4.4} & \rev{100.0 $\pm$ 0.0} & \rev{14.4 $\pm$ 4.2} & \rev{44.6 $\pm$ 7.1} \\
 & PCE & \rev{\textbf{99.4} $\pm$ 1.0} & \rev{\textbf{98.9} $\pm$ 1.0} & \rev{\textbf{97.2} $\pm$ 1.0} & \rev{100.0 $\pm$ 0.0} & \rev{\textbf{0.6} $\pm$ 1.0} & \rev{\textbf{72.9} $\pm$ 12.0} \\
\bottomrule
\end{tabular*}
\end{table}

\begin{table}[t]
\caption{Parent-only distillation ablation on Polyglot Held-in 60 (\%). \rev{Best is the highest search score in the main run, and the other columns summarize five evaluations of each method's selected agent.}}
\label{tab:parentblock}
\centering
\setlength{\tabcolsep}{8pt}
\begin{tabular}{lcccc}
\toprule
\textbf{Method} & \textbf{Best} & \rev{\textbf{Repeat mean}} & \textbf{Intersection} & \textbf{Union} \\
\midrule
GEA & 90.0 & 92.0 $\pm$ 2.5 & 85.0 & 100.0 \\
Parent distillation & 91.7 & 92.0 $\pm$ 4.3 & 78.3 & 98.3 \\
Population distillation (PCE) & \textbf{100.0} & \textbf{96.0} $\pm$ 2.8 & \textbf{88.3} & 100.0 \\
\bottomrule
\end{tabular}
\end{table}

\subsection{Performance Gains and Token Cost}
\label{sec:res:cost}

\noindent\textbf{RQ4: Does PCE achieve better performance at the cost of higher token consumption?}

\textbf{No. PCE achieves higher performance while using substantially fewer tokens.}
Table~\ref{tab:cost} shows that PCE reduces total token consumption by 36\% on SWE-bench and 59\% on Polyglot relative to GEA. These totals cover the complete evolution run, including task execution, diagnosis, self-modification, and population distillation. PCE also makes fewer model requests on both benchmarks. Its performance gains therefore do not come from a larger inference budget.

The reduction is not a result of evaluating fewer task episodes. PCE evaluates more episodes than GEA on both benchmarks, but produces substantially shorter trajectories for both solved and unsolved attempts. Extremely long trajectories also become less frequent. On SWE-bench, the number of trajectories exceeding 500,000 characters falls from 158 to 45. PCE therefore performs more evaluations while consuming fewer tokens overall. \rev{Appendix~\ref{app:tokens} provides a detailed analysis of the tokens spent in each stage and of where the savings arise.}

These results suggest that population-wide evidence makes the evolutionary process more directed. Population distillation identifies recurring outcomes and informs harness changes that can affect later task attempts. Figure~\ref{fig:motivation-intro}c provides one example, where a recurring empty-patch pattern leads to an explicit patch check and bounded repair procedure. This case does not by itself explain the aggregate savings, but it illustrates how population-guided modifications can prevent inefficient behavior from recurring. PCE therefore improves both performance and token efficiency, rather than exchanging a larger inference budget for stronger agents.

\begin{table}[t]
\caption{Performance and resource use in the main evolution runs. Token and request counts are provider totals for the complete run, covering task execution, diagnosis, self-modification, and population distillation. Distillation calls report the number made by the distillation agent. An evaluation episode is one task attempt, and trajectory length is measured in characters.}
\label{tab:cost}
\centering
\setlength{\tabcolsep}{5pt}
\begin{tabular}{lrrrr}
\toprule
\multirow{2}{*}{\textbf{Metric}}
 & \multicolumn{2}{c}{\textbf{SWE-bench}}
 & \multicolumn{2}{c}{\textbf{Polyglot}} \\
\cmidrule(lr){2-3}\cmidrule(lr){4-5}
 & \textbf{GEA} & \textbf{PCE}
 & \textbf{GEA} & \textbf{PCE} \\
\midrule
Overall success (\%)$\uparrow$ & 68.9 & \textbf{72.1} & 90.8 & \textbf{96.7} \\
Total tokens (billions)$\downarrow$ & 2.98 & \textbf{1.92} & 0.99 & \textbf{0.41} \\
Total requests$\downarrow$ & 114,629 & \textbf{69,539} & 52,034 & \textbf{28,823} \\
\midrule
Distillation calls & 0 & 758 & 0 & 775 \\
\midrule
Evaluation episodes & 1,100 & 1,200 & 1,147 & 1,177 \\
Mean trajectory length, solved (k chars)$\downarrow$ & 215 & \textbf{146} & 101 & \textbf{65} \\
Mean trajectory length, unsolved (k chars)$\downarrow$ & 354 & \textbf{242} & 126 & \textbf{60} \\
Episodes above 500k chars$\downarrow$ & 158 & \textbf{45} & 4 & \textbf{1} \\
\bottomrule
\end{tabular}
\end{table}

\begin{revblock}
\subsection{Generality Across Settings}
\label{sec:res:generality}

\noindent\textbf{RQ5: Does the advantage of PCE hold across different settings?}

\textbf{Yes. The advantage of PCE persists when evolution runs twice as long.}
As Figure~\ref{fig:rq5}b shows, we extend one Polyglot run of each method to twenty generations, twice the main schedule, and keep the other settings fixed. PCE reaches a perfect search score at generation~2. GEA still finds its best agent at generation~13, eleven generations later, and that agent reaches 91.7\%. The best agents of PCE all descend from a single early lineage. This lineage yields an agent with a perfect score on four occasions, the last at generation~11, when the population has 23 members. Appendix~\ref{app:lineage} traces these lineages in every run. PCE therefore converges faster and keeps its lead over twice as many generations, and population distillation keeps producing strong agents as the population grows.

\textbf{Yes. The harness evolved by PCE also improves agents under other execution models.}
We run the PCE agent evolved with DeepSeek Flash under seven other execution models from five providers, without further evolution, as Figure~\ref{fig:rq5}c shows. This setting examines whether the evolved harness transfers beyond the model used during evolution. The PCE agent outperforms the initial agent under every model, by 25.7 percentage points on average. Under Claude Haiku 5.5, it solves all 60 tasks, above its mean of 96.0\% under DeepSeek Flash, although the initial agent starts lower with Haiku, at 66.7\% against 72.3\%. The improvements that PCE finds from population-wide evidence therefore lie in the harness and carry over to models it never uses during evolution.
\end{revblock}

\section{Conclusion}
\label{sec:conclusion}
Agent evolution can produce useful experience that no selected parent captures alone. The gap between population coverage and best-agent performance makes this overlooked potential visible. We formulate the \emph{view from nowhere} as a principle for using the population not only to select parents, but also to understand how an individual agent should improve. PCE realizes this principle by distilling outcomes and modification histories across evolved variants into evidence for parent-specific diagnosis. Across SWE-bench Verified and Polyglot, PCE improves performance, retains its gains on held-out tasks, raises the quality of several leading variants, and brings the best agent closer to what the population collectively solves. The Polyglot ablation further shows that these gains do not arise from additional parent-only analysis. Notably, PCE also moves beyond the usual performance-cost tradeoff. It produces stronger agents while using fewer total tokens, rather than purchasing improvement with a larger inference budget. The broader lesson is that an evolutionary population can serve as a source of knowledge, not merely a pool of candidates. We envision that this perspective motivates further study of how population evidence can be attributed, compressed, and transferred as evolutionary histories grow and self-improving systems move beyond coding agents.

\bibliographystyle{bibstyle}
\bibliography{references}

\clearpage
\appendix
\section{Experimental Setup}
\label{app:details}

\paragraph{Models and provider.}
Task execution and self-modification use DeepSeek-V4-Flash on SWE-bench and DeepSeek-V4.1-Flash on Polyglot. Diagnosis and population distillation use DeepSeek-V4-Pro in all runs. Models are accessed through OpenRouter with the provider pinned to CoreWeave. \rev{The transfer evaluations in Figure~\ref{fig:rq5}c run the evolved agents unchanged through ZenMux with seven other execution models, namely Claude Haiku 5.5, Gemini 3.5 Flash Lite, GPT-6 Luna, GPT-5.6 Luna, GPT-5.4 Nano, MiniMax M3, and GLM-5.3 Flash. Thinking is disabled as in the evolution runs, except for Gemini 3.5 Flash Lite, which always reasons. Figure~\ref{fig:rq5}c abbreviates Flash Lite as FL.} 

\begin{table}[h]
\caption{Search scores of the children in all reported \rev{SWE-bench} runs. Each cell lists the two children of a generation, scored out of 60 tasks. An entry of 0/10 denotes an evaluation stopped at the ten-task gate, and a dash denotes self-improvement that produced no patch. Min, Max, and Median are over fully evaluated children. \rev{Bold marks the best agents of each run and their lineage through the inheriting parents. Run~1 is the main run.}}
\label{tab:runs}
\centering
\small
\setlength{\tabcolsep}{3pt}
\resizebox{\linewidth}{!}{%
\begin{tabular}{lccccccccccccc}
\toprule
& \multicolumn{10}{c}{\textbf{Generation}} & \multicolumn{3}{c}{\textbf{Fully evaluated}} \\
\cmidrule(lr){2-11}\cmidrule(lr){12-14}
\textbf{Run} & 0 & 1 & 2 & 3 & 4 & 5 & 6 & 7 & 8 & 9 & $n$ & Min/Max & Median \\
\midrule
PCE 1 & \textbf{40}, 34 & 38, 40 & 40, 40 & 42, 37 & 44, 38 & \textbf{43}, 42 & 37, 43 & 39, 42 & \textbf{45}, \textbf{45} & 42, 42 & 20 & 34 / 45 & 41.0 \\
PCE 2 & \textbf{40}, 29 & \textbf{40}, 36 & 42, \textbf{45} & 42, 40 & 41, 40 & 39, 40 & 39, 43 & 34, 43 & 39, 39 & 38, \textbf{45} & 20 & 29 / 45 & 40.0 \\
GEA 1 & \textbf{41}, 31 & 0/10, 33 & \textbf{42}, 39 & 40, 37 & 39, 19 & 38, 39 & 41, 38 & 39, 39 & 0/10, \textbf{43} & 39, 41 & 18 & 19 / 43 & 39.0 \\
GEA 2 & \textbf{25}, 15 & \textbf{27}, 19 & \textbf{28}, 25 & \textbf{39}, 13 & 37, 27 & \textbf{40}, 21 & \textbf{39}, 30 & 39, \textbf{39} & 35, 37 & 37, \textbf{41} & 20 & 13 / 41 & 32.5 \\
\bottomrule
\end{tabular}}
\end{table}

\begin{table}[h]
\caption{\rev{Search scores of the children in all reported Polyglot runs, in the format of Table~\ref{tab:runs}. Run~1 is the main run, and Run~3 lists the first ten generations of a twenty-generation run. The rows below the rule are the single-run ablations in Figure~\ref{fig:rq5}a.}}
\label{tab:runs-polyglot}
\centering
\small
\setlength{\tabcolsep}{3pt}
\resizebox{\linewidth}{!}{%
\begin{tabular}{lccccccccccccc}
\toprule
& \multicolumn{10}{c}{\textbf{Generation}} & \multicolumn{3}{c}{\textbf{Fully evaluated}} \\
\cmidrule(lr){2-11}\cmidrule(lr){12-14}
\textbf{Run} & 0 & 1 & 2 & 3 & 4 & 5 & 6 & 7 & 8 & 9 & $n$ & Min/Max & Median \\
\midrule
PCE \rev{1} & 38, \textbf{47} & 38, 45 & 42, \textbf{58} & 58, 59 & \textbf{60}, 57 & 45, 41 & 57, 36 & 58, 48 & 49, 34 & 59, 49 & 20 & 34 / 60 & 48.5 \\
\rev{PCE 2} & \rev{\textbf{40}, 37} & \rev{56, \textbf{42}} & \rev{58, 56} & \rev{54, 54} & \rev{\textbf{59}, 57} & \rev{57, 53} & \rev{58, 56} & \rev{\textbf{59}, 53} & \rev{\textbf{59}, 58} & \rev{\textbf{59}, 56} & \rev{20} & \rev{37 / 59} & \rev{56.0} \\
\rev{PCE 3} & \rev{\textbf{41}, 38} & \rev{\textbf{43}, 48} & \rev{\textbf{60}, 55} & \rev{57, 57} & \rev{\textbf{60}, \textbf{60}} & \rev{51, 39} & \rev{59, 59} & \rev{53, 57} & \rev{56, 55} & \rev{57, 55} & \rev{20} & \rev{38 / 60} & \rev{55.5} \\
GEA \rev{1} & 42, \textbf{47} & \textbf{50}, 41 & 51, 50 & 47, \textbf{54} & 49, 44 & 47, 50 & 53, 48 & 44, 51 & 47, 47 & 49, 41 & 20 & 41 / 54 & 47.5 \\
\rev{GEA 2} & \rev{\textbf{45}, 43} & \rev{37, 39} & \rev{43, 42} & \rev{49, 33} & \rev{37, 32} & \rev{\textbf{45}, 34} & \rev{46, \textbf{46}} & \rev{37, 48} & \rev{\textbf{51}, 47} & \rev{43, 45} & \rev{20} & \rev{32 / 51} & \rev{43.0} \\
\rev{GEA 3} & \rev{\textbf{40}, 39} & \rev{39, 42} & \rev{\textbf{41}, 40} & \rev{47, 43} & \rev{45, 38} & \rev{41, 38} & \rev{38, 37} & \rev{\textbf{49}, 42} & \rev{\textbf{49}, 47} & \rev{38, 36} & \rev{20} & \rev{36 / 49} & \rev{40.5} \\
\midrule
Parent dist. & \textbf{45}, 48 & 45, 46 & 47, \textbf{47} & \textbf{45}, 51 & 0/10, 44 & 45, 50 & 44, 52 & --, 54 & 48, 54 & 53, \textbf{55} & 18 & 44 / 55 & 47.5 \\
\rev{Member summ.} & \rev{40, \textbf{40}} & \rev{41, 50} & \rev{0/10, 46} & \rev{55, 51} & \rev{51, 55} & \rev{\textbf{50}, 46} & \rev{50, 49} & \rev{46, 46} & \rev{50, 55} & \rev{\textbf{56}, 48} & \rev{19} & \rev{40 / 56} & \rev{50.0} \\
\rev{Evidence} & \rev{\textbf{38}, 37} & \rev{\textbf{49}, 55} & \rev{\textbf{59}, 54} & \rev{\textbf{58}, 50} & \rev{\textbf{59}, 57} & \rev{0/10, 41} & \rev{42, 36} & \rev{34, 41} & \rev{35, 40} & \rev{--, --} & \rev{17} & \rev{34 / 59} & \rev{42.0} \\
\rev{Observation} & \rev{\textbf{37}, \textbf{40}} & \rev{\textbf{57}, 36} & \rev{\textbf{56}, 52} & \rev{\textbf{44}, \textbf{56}} & \rev{55, 57} & \rev{\textbf{59}, 36} & \rev{\textbf{59}, \textbf{39}} & \rev{\textbf{37}, 39} & \rev{\textbf{59}, 41} & \rev{\textbf{59}, 40} & \rev{20} & \rev{36 / 59} & \rev{48.0} \\
\rev{Evid. + obs.} & \rev{\textbf{41}, \textbf{36}} & \rev{25, \textbf{41}} & \rev{\textbf{58}, \textbf{58}} & \rev{\textbf{60}, \textbf{60}} & \rev{59, 58} & \rev{59, 59} & \rev{\textbf{60}, 58} & \rev{59, 59} & \rev{\textbf{60}, 58} & \rev{59, 59} & \rev{20} & \rev{25 / 60} & \rev{59.0} \\
\bottomrule
\end{tabular}}
\end{table}

\paragraph{Task sets.}
Both benchmarks use DGM's 60-task subsets as Held-in 60 \citep{zhang2025dgm}. Each subset is partitioned into ten gate tasks and fifty further tasks, and a child passes the gate with at least one solved gate task on SWE-bench and at least four on Polyglot. The SWE-bench Held-in 60 comprises 35 Django and 25 Sphinx tasks. SWE-bench Held-out 140 comprises DGM's additional 118 Django and 22 Astropy tasks, and the separate SWE-bench Held-out 60 spans 12 repositories. Polyglot Held-out 165 is the remainder of the 225 tasks. The best agent of \rev{the main run of each method} is selected by its Held-in 60 search score and evaluated without further self-modification. It is evaluated five times on Held-in 60 and on SWE-bench Held-out 60, and once on Held-out 140 and Held-out 165. Overall weights the mean held-in success rate and the single held-out success rate by their task counts.

\paragraph{Additional SWE-bench Held-out 60.}
We construct this set by first validating SWE-bench Verified instances in our evaluation environment.
We sample 60 tasks using fixed seed, requiring representation from every repository in the candidate pool and matching the held-in difficulty distribution as closely as possible.
The resulting set spans 12 repositories.
We evaluate the initial agent and the best agents evolved by GEA and PCE \rev{in the main run} five times on this set.

\paragraph{Evolution loop.}
Each run spans ten generations with two children per generation. \rev{One Polyglot run of each method is extended to twenty generations. The maximum generation only bounds the loop, so its first ten generations follow the same procedure as a ten-generation run.}
Generation~0 uses single-parent diagnosis from the initial harness, with a separately sampled task record for each child.
Later generations select two parents ($k=2$) using GEA's performance-novelty rule and uniformly sample one eligible task per parent.
Each diagnosis receives both records, with the inheriting parent's record first.
PCE additionally supplies a population block for that parent's task when available.
Modifications are restricted to \texttt{coding\_agent.py} (episode logic), \texttt{harness\_def.py} (instructions and truncation handling), \texttt{tools/}, and \texttt{utils/}, excluding evaluation and orchestration helpers.
The model, output cap, and tool loop remain fixed.
Children producing harness patches are evaluated on ten gate tasks and, if they pass, the remaining fifty.
Archive admission requires passing the gate, producing at least one non-empty task patch, and restricting harness changes to the permitted paths.
The archive is never pruned.
All evaluated children enter $\mathcal{V}_{t+1}$, regardless of archive admission, making their records available for subsequent distillation.

\paragraph{Population distillation.}
The distillation agent runs with a budget of 250 steps and returns an observation of at most 60 words and zero to three candidate files. Program-computed outcome counts are rendered with the observation. Distillation is skipped when fewer than three records exist on the task, so the first opportunity is generation~1, and the diagnosis proceeds without a block when no usable observation is returned.

\begin{table}[t]
\caption{Proposal counts from five diagnoses per condition. I: instructions. E: episode logic. T: new tool. Dashes indicate untested trajectory controls due to insufficient eligible trajectories at generation 1.}
\label{tab:fixedcontext}
\centering
\setlength{\tabcolsep}{8pt}
\begin{tabular}{lccc}
\toprule
\multirow{2}{*}{\textbf{Prompt condition}}
 & \textbf{Sphinx-7748}
 & \textbf{Sphinx-8638}
 & \textbf{Sphinx-9229} \\
 & I / E / T & I / E / T & I / E / T \\
\midrule
GEA & 3 / 2 / 0 & 4 / 0 / 1 & 5 / 0 / 0 \\
+Success & --- & 4 / 0 / 1 & 5 / 0 / 0 \\
+Own failures & --- & 4 / 0 / 1 & 5 / 0 / 0 \\
+Population failures & --- & 3 / 0 / 2 & 5 / 0 / 0 \\
+Block & 0 / \textbf{5} / 0 & 5 / 0 / 0 & 3 / \textbf{2} / 0 \\
\bottomrule
\end{tabular}
\end{table}

\paragraph{Published comparison settings.}
Table~\ref{tab:final} lists task-execution models and generation counts. GEA and PCE produce two children per generation, compared with one for DGM. Haiku/Sonnet denotes GEA's Haiku 4.5 to Sonnet 4.5 schedule. DGM's model-transfer entries reuse the evolved harness without additional search.

\begin{table}[h]
\caption{Repeated and held-out evaluations of the \rev{selected} agents. Rep 1--5 are solved-task counts, and the last column is mean success percentage $\pm$ sample standard deviation. Held-out 140 and Held-out 165 are evaluated once, and their rows list the solved counts and success percentages of the initial agent, GEA, and PCE. All tasks of a set remain in the denominator, including never-graded attempts.}
\label{tab:repeats}

\centering
\small
\begin{tabular}{lllcc}
\toprule
\textbf{Benchmark} & \textbf{Set} & \textbf{Agent} & \textbf{Rep 1--5} & \textbf{Mean $\pm$ sd (\%)} \\
\midrule
\multirow{7}{*}{SWE-bench} & \multirow{3}{*}{Held-in 60} & Initial & 33, 33, 36, 33, 35 & 56.7 $\pm$ 2.4 \\
 & & GEA & 39, 39, 39, 42, 40 & 66.3 $\pm$ 2.2 \\
 & & PCE & 39, 43, 45, 44, 40 & 70.3 $\pm$ 4.3 \\
\cmidrule(lr){2-5}
 & \multirow{3}{*}{Held-out 60} & Initial & 38, 30, 31, 29, 32 & 53.3 $\pm$ 5.9 \\
 & & GEA & 41, 39, 39, 40, 37 & 65.3 $\pm$ 2.5 \\
 & & PCE & 40, 42, 43, 42, 44 & 70.3 $\pm$ 2.5 \\
\cmidrule(lr){2-5}
 & Held-out 140 & Initial / GEA / PCE & 70 / 98 / 102 & 50.0 / 70.0 / 72.9 \\
\midrule
\multirow{5}{*}{Polyglot} & \multirow{4}{*}{Held-in 60} & Initial & 46, 43, 45, 40, 43 & 72.3 $\pm$ 3.8 \\
 & & GEA & 55, 57, 56, 53, 55 & 92.0 $\pm$ 2.5 \\
 & & Parent distillation & 51, 55, 56, 56, 58 & 92.0 $\pm$ 4.3 \\
 & & PCE & 56, 56, 60, 58, 58 & 96.0 $\pm$ 2.8 \\
\cmidrule(lr){2-5}
 & Held-out 165 & Initial / GEA / PCE & 103 / 149 / 160 & 62.4 / 90.3 / 97.0 \\
\bottomrule
\end{tabular}
\end{table}

\section{Diagnostic Proposals under Fixed Contexts}
\label{app:fixedcontext}
We use three diagnostic contexts from a completed SWE-bench GEA run, keeping the focal parent, source code, modification history, and task record fixed.
GEA retains the paired parent's original entry; +Block adds a fixed population block.
The remaining conditions replace that entry with the paired parent's successful trajectory on another task (+Success), three of its failures on other tasks (+Own failures), or three other agents' failures on the focal task (+Population failures).
Sphinx-7748 occurs at generation~1 and lacks sufficient eligible trajectories for these replacements.
Table~\ref{tab:fixedcontext} reports five proposals per condition, each classified by its main target as instructions (I), episode logic (E), or a new tool (T).
\begin{revblock}
Adding a population block changes the main target of diagnosis in two of three contexts. On the Sphinx-7748 task, proposals addressing truncation recovery increase from two of five to all five. On the Sphinx-9229 task, the population block produces two proposals for a post-episode patch check, while GEA and the raw-trajectory controls produce only instruction-level changes. No comparable shift appears on Sphinx-8638. A population block therefore redirects diagnosis from local instruction refinement toward changes in the agent's execution logic, and distilling cross-agent patterns provides guidance that additional raw trajectories do not reproduce. Section~\ref{sec:res:process} shows the effect of this guidance on evolved agents.
\end{revblock}

\section{Additional Results}
\label{app:results}

For SWE-bench, Table~\ref{tab:runs} reports per-child search scores from two independent runs of each method, including gate outcomes and summary statistics for fully evaluated children.
Run~1 supplies the main performance and cost results, while Table~\ref{tab:cohort} summarizes both runs.
Table~\ref{tab:repeats} provides five repeated evaluations on Held-in 60 and Held-out 60, together with the single evaluation on Held-out 140.

For Polyglot, Table~\ref{tab:runs-polyglot} reports per-child search scores for GEA, parent distillation, and PCE\rev{, with three independent runs of GEA and PCE. Run~1 is the main run and supplies the performance and cost results. Runs~2 and~3 are additional independent runs, and Run~3 lists the first ten generations of a twenty-generation run. Table~\ref{tab:cohort} summarizes all three runs}.
Table~\ref{tab:repeats} lists five Held-in 60 evaluations of the initial and best evolved agents \rev{of Run~1}, including the parent-distillation control.
It also reports Held-out 165 results for the initial agent, GEA, and PCE.
These results support the main performance comparison and the population-distillation ablation.

\section{\rev{Population Analyses}}
\label{app:population}
\rev{The consolidation rate in Table~\ref{tab:cohort} measures how often children acquire successes that earlier population members find. This section defines the rate and follows the lineages through which these successes produce the best agents.}

\subsection{\rev{Consolidation rate}}
\begin{revblock}
\label{app:consolidation}
The consolidation rate measures how often a child solves a task that both of its selected parents fail but an earlier population member solves. Let $S(h)$ be the set of Held-in 60 tasks that agent $h$ solves in its search evaluation, and let $S(H)=\bigcup_{h\in H}S(h)$ for a set of agents $H$. Consider a child $h_c$ produced in generation $t\geq 1$ from the selected parents $P_t$ of Algorithm~\ref{alg:loop}, and let $\mathcal{V}_t$ be the population before the children of generation $t$ are added. The target set of $h_c$ contains the tasks that another member of $\mathcal{V}_t$ solves and that no selected parent solves,
\begin{equation*}
F_c = S\big(\mathcal{V}_t \setminus (P_t \cup \{h_0\})\big) \setminus S(P_t).
\end{equation*}
The consolidation rate of a run pools the counts over all its children,
\begin{equation*}
\mathrm{CR} = \frac{\sum_c |S(h_c) \cap F_c|}{\sum_c |F_c|}.
\end{equation*}
Pooling gives every target task the same weight. We include the children evaluated on all 60 tasks and, as with Union in Table~\ref{tab:cohort}, leave out the initial agent $h_0$. On Polyglot, every PCE run exceeds every GEA run, with rates of 59.1, 81.0, and 78.7\% for PCE against 37.0, 45.5, and 51.2\% for GEA. On SWE-bench, the two runs give 41.9 and 29.2\% for PCE and 25.5 and 40.6\% for GEA, with a higher mean for PCE.
\end{revblock}

\subsection{\rev{Lineages of the best agents on Polyglot}}
\begin{revblock}
\label{app:lineage}
The best agents of PCE come from a lineage that forms early, and PCE repeatedly evolves its best agents from this lineage. Table~\ref{tab:lineage} traces each best agent back through its inheriting parent. In all three PCE runs, the lineage of the best agents is in place by generation~2, when each task has four or five records. PCE already scores between 96.7\% and 100\% at that point, so population distillation is effective while the population is still small. The lineage is also reused. In Runs~2 and~3, one generation-1 agent, solving 42 and 43 tasks respectively, is the parent of almost every later child. PCE evolves its best agents from it, four times in Run~2 and three times in Run~3. Parent selection also returns repeatedly to one agent in GEA, so reuse alone does not explain the difference. In the main GEA run, a generation-1 agent is the parent of 14 of the 16 later children, yet its best child solves 54 tasks. In the other two GEA runs, the best agent appears at generation 8 or 13. The difference therefore lies in how diagnosis improves the reused parent, and in PCE this step is guided by population-level evidence. Figure~\ref{fig:rq5}b plots these lineages for Run~3 and the corresponding GEA run, following the higher-scoring parent when both parents of a child come from the same generation.

\paragraph{Parents of the largest gains.} The largest gains of PCE start from ordinary members. Over the first ten generations of Runs~1, 2, and~3, the largest gain of a child over its inheriting parent is 17, 19, and 18 tasks for PCE, against 6, 6, and 8 for GEA. The inheriting parents of these three PCE children solve 42, 37, and 41 tasks, below the median of the population at that time. Together with the reuse of one lineage described above, this shows that PCE does not rely on a diverse population or on a member with a rare skill. Population-level evidence lets diagnosis lift an ordinary parent in a single large step.

\begin{table}[h]
\caption{\rev{Parent of the best agents in each Polyglot run, given as generation and solved tasks, with the number of later children it parents in either parent role. The Best agents column lists their generations. Run~3 covers twenty generations.}}
\label{tab:lineage}
\centering
\revcolor
\small
\begin{tabular}{llcccc}
\toprule
\textbf{Method} & \textbf{Run} & \textbf{Parent} & \textbf{Its later children} & \textbf{Best agents} & \textbf{Solved} \\
\midrule
PCE & 1 & 2, 58 & 4 of 14 & 4 & 60 \\
PCE & 2 & 1, 42 & 16 of 16 & 4, 7, 8, 9 & 59 \\
PCE & 3 & 1, 43 & 30 of 36 & 2, 4, 4, 11 & 60 \\
\midrule
GEA & 1 & 1, 50 & 14 of 16 & 3 & 54 \\
GEA & 2 & 6, 46 & 4 of 6 & 8 & 51 \\
GEA & 3 & 12, 44 & 4 of 14 & 13 & 55 \\
\bottomrule
\end{tabular}
\end{table}
\end{revblock}

\section{\rev{Token Analysis}}
\label{app:tokens}
\begin{revblock}
This section shows where the tokens of PCE go and where its savings come from. It uses the per-call usage records of the twenty-generation Polyglot runs of Figure~\ref{fig:rq5}b, which cover every stage, while Table~\ref{tab:cost} reports provider totals for the main run. Tokens count cached and uncached input together with output.

\paragraph{Overall use.}
PCE uses a quarter as many tokens as GEA. Over twenty generations, PCE spends 698.1M tokens against 2,786.0M for GEA, and over the first ten generations the ratio is 31\%. Table~\ref{tab:tokens} lists every stage.

\paragraph{Source of the savings.}
The savings come almost entirely from task execution. Task execution saves 2,037.0M tokens and self-modification saves 208.5M, while population distillation adds 160.1M, giving a net saving of 2,087.9M. PCE evaluates more episodes than GEA, 2,400 against 2,100, yet each episode is far shorter for solved and unsolved tasks alike. Episodes above 1M tokens fall from 611 to 94, as Table~\ref{tab:episodes} shows.

\paragraph{Cost of population distillation.}
Population distillation is cheap relative to these savings. It uses 160.1M tokens, 7.7\% of the tokens that PCE saves, and GEA has no such stage.

\paragraph{Growth with population size.}
This cost stays flat as the population grows. Each distillation runs under a fixed budget of 250 steps. From generation~3 to generation~19 the population grows from 7 to 39 members. Over these generations, the distillation tokens per child stay between 3.2M and 4.8M, except for 6.9M and 7.7M in generations~12 and~17. The additional analysis of PCE therefore costs little, and its savings come from more efficient task execution.

\begin{table}[h]
\caption{\rev{Tokens by stage in the twenty-generation Polyglot runs.}}
\label{tab:tokens}
\centering
\revcolor
\small
\begin{tabular}{lcccc}
\toprule
& \multicolumn{2}{c}{\textbf{Generations 0 to 9}} & \multicolumn{2}{c}{\textbf{Generations 0 to 19}} \\
\cmidrule(lr){2-3}\cmidrule(lr){4-5}
\textbf{Stage (millions of tokens)} & GEA & PCE & GEA & PCE \\
\midrule
Task execution & 1,011.8 & 250.7 & 2,548.2 & 511.2 \\
Self-modification & 57.6 & 13.8 & 232.7 & 24.2 \\
Diagnosis & 1.8 & 1.3 & 5.0 & 2.5 \\
Population distillation & 0.0 & 64.5 & 0.0 & 160.1 \\
\midrule
Total & 1,071.2 & 330.2 & 2,786.0 & 698.1 \\
\bottomrule
\end{tabular}
\end{table}

\begin{table}[h]
\caption{\rev{Task-execution episodes in the twenty-generation Polyglot runs. Above 1M counts the episodes that use more than one million tokens.}}
\label{tab:episodes}
\centering
\revcolor
\small
\begin{tabular}{llccccc}
\toprule
& & & \multicolumn{3}{c}{\textbf{Tokens per episode (thousands)}} & \\
\cmidrule(lr){4-6}
\textbf{Generations} & \textbf{Method} & \textbf{Episodes} & \textbf{Solved mean} & \textbf{Unsolved mean} & \textbf{Median} & \textbf{Above 1M} \\
\midrule
0 to 9 & GEA & 1,200 & 764 & 1,019 & 341 & 279 \\
 & PCE & 1,200 & 186 & 382 & 70 & 43 \\
\midrule
0 to 19 & GEA & 2,100 & 1,058 & 1,598 & 417 & 611 \\
 & PCE & 2,400 & 186 & 351 & 77 & 94 \\
\bottomrule
\end{tabular}
\end{table}
\end{revblock}

\end{document}